\documentclass[10pt,twocolumn,letterpaper]{article}

\usepackage{iccv}
\usepackage{times}
\usepackage{epsfig}
\usepackage{graphicx}
\usepackage{amsmath}
\usepackage{amssymb}
\usepackage{algorithm}
\usepackage{algpseudocode}

\usepackage{graphicx}
\usepackage{amsmath}
\usepackage{amssymb}
\usepackage{booktabs}
\usepackage{tabularx}
\usepackage{pifont}
\usepackage{paralist}
\usepackage{multirow}
\usepackage{adjustbox}
\newcommand{\xmark}{\textcolor{red}{\ding{55}}}
\newcommand{\cmark}{\textcolor{blue}{\ding{51}}}
\usepackage{makecell}
\usepackage{colortbl}
\usepackage[table]{xcolor}
\definecolor{lightgray}{gray}{0.9}
\definecolor{LightCyan}{rgb}{0.88,1,1}

\usepackage[pagebackref=true,breaklinks=true,letterpaper=true,colorlinks,citecolor=blue,linkcolor=blue,bookmarks=false]{hyperref}

\usepackage[capitalize]{cleveref}
\crefname{section}{Sec.}{Secs.}
\Crefname{section}{Section}{Sections}
\Crefname{table}{Table}{Tables}
\crefname{table}{Tab.}{Tabs.}

\iccvfinalcopy 

\def\iccvPaperID{9694} 

\ificcvfinal\fi

\begin{document}

\title{CLR: Channel-wise Lightweight Reprogramming for Continual Learning}

\author{Yunhao Ge$^{1\dagger}$, Yuecheng Li$^{1*}$, Shuo Ni$^{1*}$, Jiaping Zhao$^{2}$ Ming-Hsuan Yang$^{2}$, Laurent Itti$^{1\dagger}$\\
$^{1}$University of Southern California \ \ \  $^{2}$Google Research \\
{\tt \small
$^{*}$Equal contribution as second author, \
$\dagger$correspondence to \{yunhaoge, itti\}@usc.edu} 
}


\maketitle
\ificcvfinal\thispagestyle{empty}\fi
\newcommand{\Revise}[1]{\textbf{\textcolor{blue}{#1}}}

\renewcommand{\algorithmicrequire}{\textbf{Input:}}
\renewcommand{\algorithmicensure}{\textbf{Output:}}

\begin{abstract} 
Continual learning aims to emulate the human ability to continually accumulate knowledge over sequential tasks. The main challenge is to maintain performance on previously learned tasks after learning new tasks, i.e., to avoid catastrophic forgetting. We propose a Channel-wise Lightweight Reprogramming (CLR) approach that helps convolutional neural networks (CNNs) overcome catastrophic forgetting during continual learning. We show that a CNN model trained on an old task (or self-supervised proxy task) could be ``reprogrammed" to solve a new task by using our proposed lightweight (very cheap) reprogramming parameter. 
With the help of CLR, we have a better stability-plasticity trade-off to solve continual learning problems: 
%
To maintain  \textbf{stability} and retain previous task ability, we use a common task-agnostic immutable part as the shared ``anchor" parameter set. 
We then add task-specific lightweight reprogramming parameters to reinterpret the outputs of the immutable parts, to enable \textbf{plasticity} and integrate new knowledge. 
To learn sequential tasks, we only train the lightweight reprogramming parameters to learn each new task. Reprogramming parameters are task-specific and exclusive to each task, which makes our method immune to catastrophic forgetting.
To minimize the parameter requirement of reprogramming to learn new tasks, we make reprogramming lightweight by only adjusting essential kernels and learning channel-wise linear mappings from anchor parameters to task-specific domain knowledge. We show that, for general CNNs, the CLR parameter increase is less than 0.6\% for any new task. 
Our method outperforms 13 state-of-the-art continual learning baselines on a new challenging sequence of 53 image classification datasets. Code and data are available at \small{\textcolor{magenta}{\url{https://github.com/gyhandy/Channel-wise-Lightweight-Reprogramming}}}.

\end{abstract}

\vspace{-5pt}
\section{Introduction}
\label{sec:introduction}
Continual Learning (CL) focuses on the problem of learning from a stream of data, where agents continually extend their acquired knowledge by sequentially learning new tasks or skills, while avoiding forgetting of previous tasks \cite{parisi2019continual}. 
In the literature, CL is also referred to as lifelong learning \cite{chaudhry2018efficient, aljundi2017expert, parisi2019continual} and sequential learning \cite{aljundi2018selfless}. This differs from standard train-and-deploy approaches, which cannot incrementally learn without catastrophic interference across tasks \cite{french1999catastrophic}. \textit{How to avoid catastrophic forgetting} is the main challenge of continual learning, which requires that the performance on previous learned tasks should not degrade significantly over time when new tasks are learned. This is also related to a general problem in neural network design, the stability-plasticity trade-off \cite{Grossberg1982StudiesOM}, where plasticity represents the ability to integrate new knowledge, and stability refers to the ability to retain previous knowledge \cite{de2021continual}. 

\begin{figure}
\centering
\includegraphics[width=\linewidth]{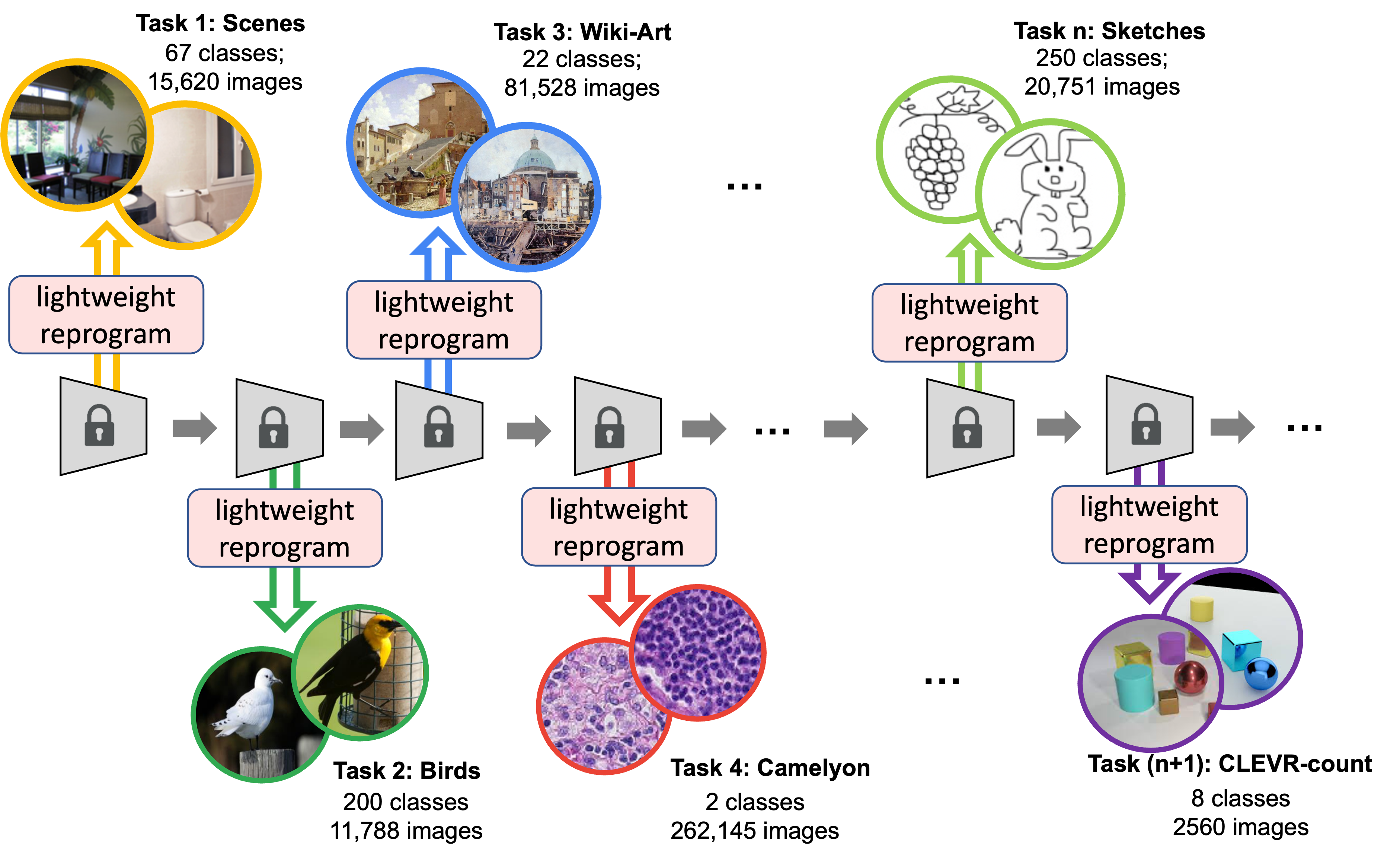}
\caption{Equipped with a task-agnostic immutable CNN model, 
our approach "reprogram" the CNN layers to each new task with lightweight task-specific parameters (less than 0.6\% of the original model) to learn sequences of disjoint tasks, assuming data from previous tasks is no longer available while learning new tasks.}
\label{fig:teaser}
\vspace{-10pt}
\end{figure}

Dynamic Network methods have been shown to be among the most successful ones to solve continual learning, which usually shows great stability on previous tasks and alleviates the influence of catastrophic forgetting by dynamically modify the
network to solve new tasks, usually by network expansion \cite{parisi2019continual, wen2021beneficial, cheung2019superposition, wortsman2020supermasks}. For stability, a desirable approach would be to fix the backbone and learn extra task-specific parameters on top of it, which will have no catastrophic forgetting. However, the number of parameters in such methods can quickly become very large. How to reduce the amount of required extra parameters is still a challenging problem. 
To solve the above issues, our approach is based on three motivations: 


\noindent (1) \textbf{Reuse instead of re-learn.} Adversarial Reprogramming \cite{elsayed2018adversarial} is a method to ``reprogram" an already trained and frozen network from its original task to solve new tasks by perturbing the input space without re-learning the network parameters. It computes a single noise pattern for each new task. This pattern is then added to inputs for the new task and fed through the original network. The original network processes the combined input + noise and generates an output, which is then remapped onto the desired new task output domain. For example, one pattern may be computed to reprogram a network originally trained on ImageNet \cite{russakovsky2014imagenet} to now solve MNIST \cite{deng2012mnist}. The same pattern, when added to an image of an MNIST digit, would trigger different outputs from the ImageNet-trained network for the different MNIST classes (e.g., digit 0 + pattern may yield ``elephant"; digit 1 + pattern ``giraffe", etc). These outputs can then be re-interpreted according to the new task (elephant means digit 0, etc). Although the computation cost is prohibitively high compared to baseline lifelong learning approaches, here we borrow the reprogramming idea; but we conduct more lightweight yet also more powerful reprogramming in the parameter space of the original model, instead of in the input space.

\noindent (2) \textbf{Channel-wise transformations may link two different kernels.} GhostNet \cite{han2020ghostnet} could generate more feature maps from cheap operations applied to existing feature maps, thereby allowing embedded devices with small memory to run effectively larger networks.
This approach is motivated by near redundancy in standard networks: after training, several learned features are quite similar. 
As such, Han {\em et al.} \cite{han2020ghostnet} generate some features as linear transformations of other features. 
Inspired by this, our approach augments a network with new, linearly transformed feature maps, which can cheaply be tailored to individual new tasks.

\noindent (3) \textbf{Lightweight parameters could shift model distribution.} BPN \cite{wen2021beneficial} adds beneficial perturbation biases in the fully connected layers to shift the network parameter distribution from one task to another, which is helpful to solve continual learning. This is cheaper than fine-tuning all the weights in the network for each task, instead tuning only one bias per neuron. Yet, this approach provided good lifelong learning results. However, the method could only handle fully connected layers and the performance is bounded by the limited ability of the bias parameters to change the network (only 1 scalar bias for each neuron). Our method instead designs more powerful reprogramming patterns (kernels) for the CNN layers, which could lead to better performance on each new task. 

\begin{figure*}[t]
\vspace{-30pt}
\begin{center}
\includegraphics[width=.85\linewidth]{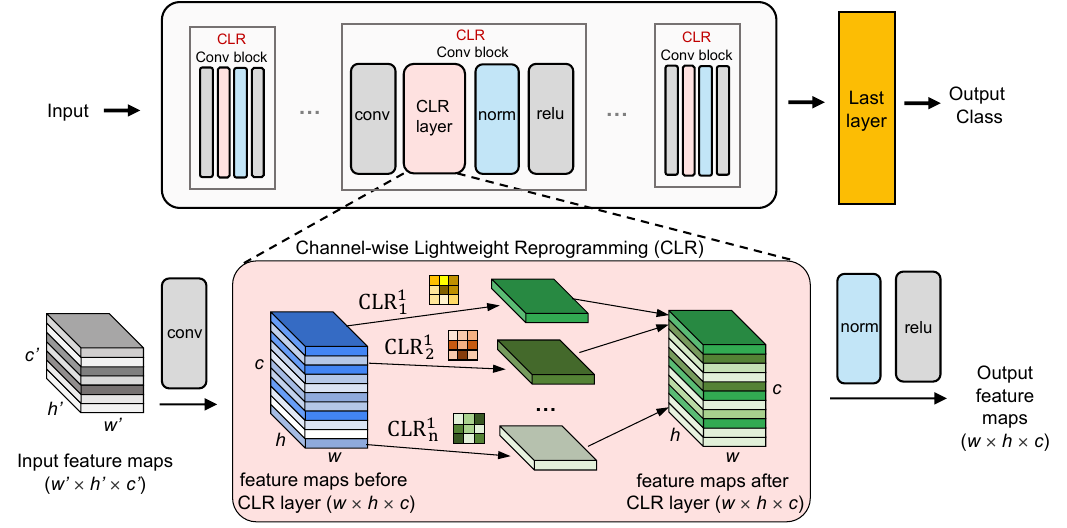}
\end{center}
  \caption{
  Proposed continual learning model with channel-wise lightweight reprogramming (CLR) layers. All gray blocks are fixed parameters. (top) General network architecture. (bottom) Details of CLR reprogramming layer: for each channel $k\in[1..c]$ of an original $w\times h\times c$ feature map (blue), a 3x3 kernel is learned to reprogram the feature towards the new task (green), without modifying the original conv parameters (grey).}
\label{fig:CLR}
\vspace{-10pt}
\end{figure*}

Drawing from these three ideas, we propose channel-wise lightweight reprogramming (CLR). We start with task-agnostic immutable parameters of a CNN model pretrained on a relatively diverse dataset (e.g., ImageNet-1k, Pascal VOC, ...) if possible, or on a self-supervised proxy task, which requires no semantic labels.  We then adaptively ``reprogram" the immutable task-agnostic CNN layers to adapt and solve new tasks by adding lightweight channel-wise linear transformation on each channel of a selected subset of Convolutional layers (Fig~\ref{fig:CLR}). The added reprogramming parameters are 3x3 2D convolutional kernels, each working on separate channels of feature maps after the original convolutional layer. CLR is very cheap but still powerful, with the intuition that different kernel filters could be reprogrammed with a task-dependent linear transformation.

\vspace{2pt}

\noindent The main contributions of this work are: 
\begin{compactitem}
    \item We propose a novel continual learning solution for CNNs, which involves reprogramming the CNN layers trained on old tasks to solve new tasks by adding lightweight task-specific reprogramming parameters. This allows a single network to learn potentially unlimited input-to-output mappings, and to switch on the fly between them at runtime.
    
    
    \item Out method achieves better stability-plasticity balance compared to other dynamic network continual learning methods: it does not suffer from catastrophic forgetting problems and requires limited extra parameters during continual learning, which is less than 0.6\% of the original parameter size for each new task. 
    
    \item Our method achieves state-of-the-art  performance on task incremental continual learning on a new challenging sequence of 53 image classification datasets. 
    
\end{compactitem}








\section{Related Work}
\vspace{-3pt}
\noindent {\bf Continual learning.} 
Continual learning methods can be broadly categorized in three approaches~\cite{de2021continual}: Replay-based, Regularization-based, and Dynamic network-based.

\underline{\textit{(1) Replay methods}} use a buffer containing sampled training data from previous tasks, as an auxiliary to a new task's training set. The buffer can be used either at the end of the task training (iCaRL \cite{rebuffi2017icarl}, ER \cite{robins1995catastrophic}) 
or during training (GSS, AGEM, AGEM-R, DER, DERPP \cite{lopez2017gradient, aljundi2019gradient, chaudhry2018efficient, buzzega2020dark}). 
Rehearsal schemes \cite{rebuffi2017icarl,  rolnick2019experience, isele2018selective, chaudhry2019tiny} explicitly retrain on a limited subset of stored samples while training on new tasks. \cite{lopez2017gradient, chaudhry2018efficient} use constrained optimization as an alternative solution leaving more leeway for backward/forward transfer. Pseudo rehearsal methods use output of previous model \cite{robins1995catastrophic} or use generates pseudo-samples with a saved generative model \cite{shin2017continual}. Rehearsal methods have the drawback that (1) they require an extra buffer to save samples (related to dataset hardness), (2) they easily overfit to the subset of stored samples during replay. (3) the performance seems to be bounded by joint training, which is influenced by the number of tasks.

\underline{\textit{(2) Regularization methods}} add an auxiliary loss term to the primary task objective to constraint weight updates. Data-focused methods (LwF\cite{li2017learning}, LFL\cite{jung2016less}, DMC \cite{zhang2020class}) use knowledge distillation from previous models trained on old tasks to constrain the model being trained on the new tasks. Prior-based methods (EWC \cite{kirkpatrick2017overcoming}, IMM \cite{lee2017overcoming}, SI \cite{zenke2017continual}, MAS \cite{aljundi2018memory}) estimate the importance of network parameters to previously learned tasks, and penalize  changing  important parameters while training  new tasks. The drawbacks of these methods are expensive computation of importance estimation and suboptimality on new tasks, especially with large numbers of tasks.

\underline{\textit{(3) Dynamic network methods}} dynamically modify the network to solve new tasks, usually by network expansion. When no constraints apply to architecture size, one can grow new branches for new tasks, while freezing previous task parameters \cite{rusu2016progressive, xu2018reinforced}, or dedicate a model copy to each task \cite{aljundi2017expert}. To save memory, recent methods keep architecture remains static, with fixed parts allocated to each task. Previous task parts are masked out during new task training, either imposed at parameters level \cite{fernando2017pathnet, mallya2018packnet}, or unit level \cite{serra2018overcoming}. SUPSUP \cite{wortsman2020supermasks}), PSP \cite{cheung2019superposition} assign a fixed set of model parameters to a task and avoid over-writing them when new tasks are learned.
Dynamic network methods have been shown to be among the most successful ones in solving continual learning. For stability, a desirable approach would be to fix the backbone and learn extra task-specific parameters on top of it, which will have no catastrophic forgetting. SUPSUP \cite{wortsman2020supermasks} uses a randomly initialized fixed backbone and learns a task-specific supermask for each new task, while the performance is bounded by the ability of the fixed backbone. CCLL \cite{singh2020calibrating}, and EFTs \cite{verma2021efficient} use fixed backbones trained on the first task, and learn task-specific group convolution parameters. The performance is sensitive to the first task, and the extra group convolution hyper-parameter is not straightforward to train.

Our channel-wise lightweight reprogramming method also belongs to dynamic network continual learning methods. Inspired by ghost networks, BPN, and Adversarial reprogramming, we reprogram the immutable task-agnostic parameters and apply lightweight (computation and memory cheap) extra parameters to learn new tasks.

\noindent{\bf Meta learning.} Meta learning aims to improve the learning algorithm itself, given the experience of multiple learning episodes \cite{hospedales2021meta}. In contrast, conventional ML improves model predictions over
multiple data instances. During meta-learning, an outer (or upper/meta) algorithm updates and improves the inner learning algorithm (e.g., one image classification task). For instance, the outer objective could be the generalization performance or learning speed of the inner algorithm. Meta-learning has the following main difference with CL: similar to transfer learning, Meta-learning focuses on finding a better learning strategy or initialization, and cares about the performance on a new or generalized task, while performance on previous tasks is not the interest. Thus, meta-learning could be used as a preliminary step in CLR, to optimize our shared initial parameter $\theta$ in CLR. Meta-learning can learn from a single task in multiple domains \cite{Thrun1998LearningTL, finn2017model} or different tasks \cite{franceschi2018bilevel, liu2018darts, andrychowicz2016learning}, which requires access to some data from all base tasks or domains at the same time, while our method and CL assume sequential learning and cannot access the data from previous tasks. Also, meta-learning assumes the learned general setting will be used in similar tasks as the base tasks, while CL does not assume that previous tasks are similar to future tasks.

\vspace{-3pt}
\section{Proposed Method}
\vspace{-3pt}
\noindent \textbf{Problem setting.}
In this paper, we focus on the task-incremental setting of continual learning, where data arrives sequentially in batches, and one batch corresponds to one task, such as a new set of classes or a new dataset to be learned. For a given continual learning algorithm, the goal is to obtain a high average performance on all previously learned tasks after learning the final task. At test time, just like PSP \cite{cheung2019superposition}, CCLL\cite{singh2020calibrating}, EFTs\cite{verma2021efficient}, EWC \cite{kirkpatrick2017overcoming}, ER\cite{robins1995catastrophic}, etc, we assume that a task oracle is provided during inference, showing which task a given sample belongs to.

\noindent \textbf{Structure.}
We first introduce how our proposed Channel-wise Lightweight Reprogramming parameters reprogram the immutable task-agnostic backbone by conducting channel-wise linear transformation after the original convolutional layer to change the original Conv-block as CLR-Conv block, and then develop a new CLR-reprogrammed network to solve new task. (Sec.~\ref{sec:3.1}). Then, we introduce how CLR-reprogrammed networks can be used to solve continual learning tasks (Sec.~\ref{sec:3.2}).

\vspace{-3pt}
\subsection{Channel-wise Lightweight Reprogramming}
\vspace{-3pt}
\label{sec:3.1}

Our proposed Channel-wise Lightweight Reprogramming method is equipped with an immutable task-agnostic backbone and creates task-specific lightweight reprogramming parameters to solve new tasks.
Here, we use a fixed backbone as a task-shared immutable structure. This  differs from SUPSUP \cite{wortsman2020supermasks}, which uses a randomly initialized fixed backbone, and CCLL \cite{singh2020calibrating}, EFTs \cite{verma2021efficient}, which use fixed backbones trained on the first task. We use a more general and compatible way with less requirement to obtain the backbone: the fixed backbone could be pretrained with supervised learning on a relatively diverse dataset (e.g., on ImageNet-1k \cite{russakovsky2014imagenet}, or Pascal VOC \cite{Everingham15}), or with self-supervised learning on proxy tasks, such as DINO \cite{caron2021emerging} and SwAV \cite{caron2020unsupervised}, which requires no semantic labels -- we will see that our approach is robust to the choice of a pretraining dataset in the Sec.~\ref{sec:4.5} experiments). This fixed backbone can provide a diverse set of visual features. However, those need to be reprogrammed later for individual tasks, which is achieved by our CLR layers. 


Specifically, we use channel-wise linear transformation to reprogram the feature map generated by an original convolutional kernel, to generate a new task-specific feature map for the new task. 

\begin{algorithm}
\caption{CLR Layers}\label{alg:alg-1}
\begin{algorithmic}
\Require

Feature map $X'_{w\times h \times c}$ (output from the original conv layer $F$), $3\times3$ 2D $CLR$ reprogram kernels $\times c$

\Ensure Reprogrammed feature map $\hat{X'}_{w\times h \times c}$

\State $p \gets \lfloor 3/2 \rfloor$
\State $X' \gets$ zero-padding($X'$, $p$)
\For{$k\in[1..c]$}

\State $\hat{X'}[k]$ $\gets$ $CLR_{k}$($X'[k]$) \quad \emph{$X'[k]$ is the $k$-th channel}
\EndFor
\end{algorithmic}
\end{algorithm}

Fig.~\ref{fig:CLR} shows the structure of the proposed CLR. CLR is compatible with any convolutional neural network (CNN).
A CNN usually consists of several Conv blocks (e.g., Residual block) (Fig.~\ref{fig:CLR} top), which contain a convolutional layer (conv), normalization layer (e.g., batch normalization), and activation layer (e.g., relu). Our method treats a pretrained CNN as a general and task-agnostic immutable parameter backbone for all future tasks, so we fix its parameters. (More details on the choice of a pretraining backbone in the Sec.~\ref{sec:4.5} experiments).  To reprogram the CNN to solve a new task in continual learning, we add lightweight trainable parameters by changing each of the original Conv blocks into a CLR-Conv block, thereby creating a CLR-reprogrammed CNN (Fig.~\ref{fig:CLR} top).
Specifically, as shown in Fig.~\ref{fig:CLR} (top), a CLR-Conv block is obtained by adding a channel-wise lightweight reprogramming layer (CLR layer) after each of the original fixed conv layers. (For parameter saving and preventing overfit, 1$\times$1 conv layers are excluded.) Each CLR layer conducts linear transformations on each channel of the original feature map after the fixed conv layer to reprogram the features. Here, the linear transformation is represented with 3x3 2D convolutional kernels conducted on single channel feature map.  Fig.~\ref{fig:CLR} (bottom) illustrates the details of Channel-wise linear transformation for reprogramming. For each convolutional kernel $f_{k}()$, given the input feature $X$, we obtain one channel of feature $x'_k = f_{k}(X)$ after the process, all Channel-wise features form the output feature map $X'$. Our Channel-wise reprogramming Linear transformation  is applied on each channel $x'_k$ of the output feature map $X'$. For each kernel $f_{k}()$, we have a corresponding reprogramming 3x3 2D kernel $CLR_k$, which takes the single channel output $x'_k$ as input and conducts a linear transformation to obtain the reprogrammed feature $\hat{x'}_k$:
\begin{equation}
    \hat{x'}_k = CLR_{k}(x'_k) = CLR_{k}(f_{k}(X))
\end{equation}
%
Algorithm~\ref{alg:alg-1} shows the pseudocode of the features before and after a CLR layer. We initialize the CLR layer as an identity transformation kernel (e.g., in a 3x3 2D kernel, the center parameter is one and all others are zero). This setting is crucial for training efficiency and performance, as it favors keeping the general feature extractors in the original fixed model, while at the same time it allows achieving adaptive reprogramming, based on the loss function for the new task. 

For CLR-reprogrammed CNNs, the original conv layers in the backbone are fixed, the trainable parameters include the CLR layer after each fixed conv layer (normalization layer is optional to train), and the last fully-connected layer. 
For CLR-reprogrammed Resnet-50 \cite{he2016deep}, the trainable CLR layer takes only 0.59\% of the parameter of the original fixed Resnet-50 backbone. This efficient parameter property makes CLR-reprogrammed networks easy to deploy for continual tasks.

\begin{figure*}[th]
\vspace{-30pt}
\begin{center}
\includegraphics[width=\linewidth]{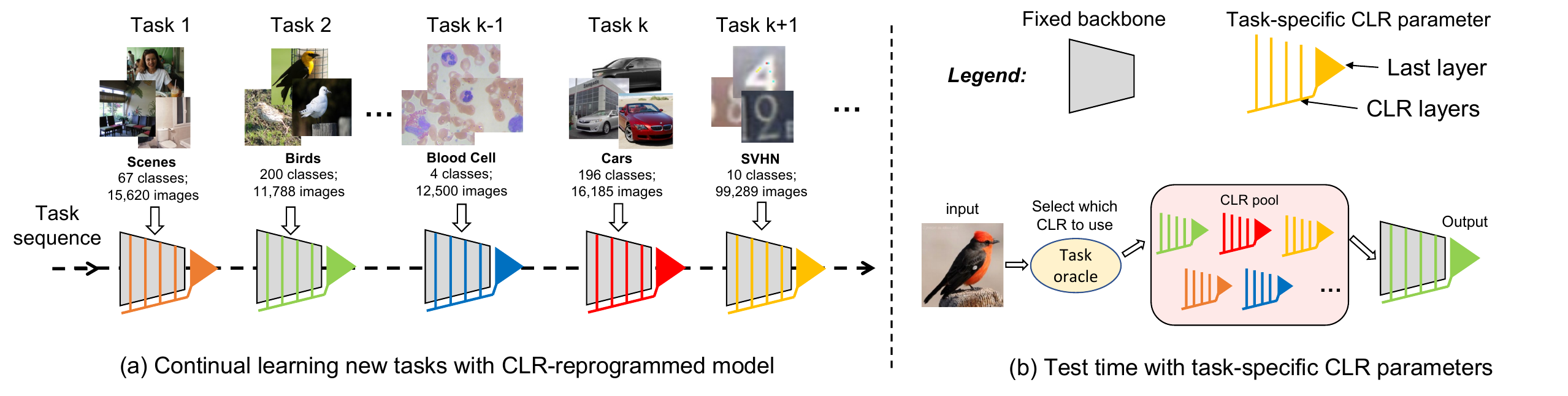}
\end{center}
  \caption{CLR-reprogrammed CNNs for continual learning. (a) In learning time, a CNN model could be reprogrammed by Channel-wise Lightweight Reprogramming parameters to solve continual new tasks. Only the CLR layers need to be trained in each reprogramming. (b) In test time, task oracle will select which task-specific CLR parameters to use and make the final decision.}
\label{fig:CLR-CL}
\vspace{-10pt}
\end{figure*}

\subsection{CLR for Continual Learning}
\vspace{-5pt}
\label{sec:3.2}


In task incremental continual learning, the model faces a sequence of tasks.
As shown in  Fig.~\ref{fig:CLR-CL}(a), during continual learning, a CLR-reprogrammed model learns each task at a time, and all tasks share the same fixed pretrained backbone.
Each task has a trainable task-specific CLR parameter set consisting of the CLR layers after each original conv layer and the last linear layer. 
During testing (Fig.~\ref{fig:CLR-CL}(b)), we assume a perfect task oracle (as assumed by our baselines shown in the next section) which tells the model which task the test image belongs to. The fixed backbone equipped with the corresponding task-specific CLR parameter  makes the final decision.

Due to the absolute parameter isolation in CLR (i.e., CLR layers are completely specific and separate for each task, and the shared backbone is not changed at all), our method's performance on every task is not influenced by increasing number of tasks (similar to SUPSUP \cite{wortsman2020supermasks}, CCLL \cite{singh2020calibrating}, and EFT \cite{verma2021efficient}). Theoretically, the CLR-reprogrammed model can learn as many tasks as needed in a continual learning setting and keep the optimal performance on each task with no accuracy decrease when the  number of tasks increase, but with a  0.59\% increase in parameters for each new task.

\vspace{-5pt}
\section{Experiments and results}
\label{sec:4}
In this section, we compare our CLR-reprogrammed model to baselines on a challenging 53-dataset with large variance (Sec.~\ref{sec:4.1}). We evaluate task accuracy fluctuation after learning new tasks to assess forgetting (Sec.~\ref{sec:4.2}). Then, we evaluate the average accuracy on all learned tasks  so far during continual learning (Sec.~\ref{sec:4.3}). We analyze the network parameters and computation cost during continual learning (Sec.~\ref{sec:4.4}). We conduct ablation study to analyze the influence of different immutable backbones (Sec.~\ref{sec:4.5})



\begin{table*}[th]
\vspace{-20pt}
\begin{center}
\small
 \rowcolors{1}{}{lightgray}
 \resizebox{\textwidth}{!}{
\begin{tabular}{c c c c c c c}
\toprule
    Comparison  & 53-dataset (ours) & 8-dataset \cite{aljundi2018memory, aljundi2017expert} & ImageNet \cite{russakovsky2014imagenet} & Fine grained 6 tasks \cite{russakovsky2014imagenet} \cite{lee2017overcoming} & Cifar100 \cite{krizhevsky2009learning}  & F-CelebA \cite{rebuffi2017learning} \\
\midrule
  \# tasks & \textbf{53} & 8 & 20 & 6 & 20 & 10  \\ 
  \# Classes & 1,584 & 738 & 1000 & \textbf{1943} & 100 & 20   \\  
  \# Images & \textbf{1,811,028} &  166,360  & 1,300,000 & 1,440,086 & 60,000 & 1,189   \\ 
  \# different classification target & \textbf{5} & 1 & 1 & 2 & 1 & 1   \\ 
  Mix image style (nature/artifact)  & \cmark & \cmark & \xmark & \cmark & \xmark & \xmark \\
  Mix super/fine-class
classification   & \cmark & \cmark & \cmark & \xmark & \xmark & \xmark \\
\bottomrule
\end{tabular}}
\end{center}
\caption{Comparison of 53-dataset with other benchmark datasets including Cifar-100 \cite{krizhevsky2009learning}, F-CelebA \cite{rebuffi2017learning}, Fine-grained 6 tasks \cite{russakovsky2014imagenet} \cite{lee2017overcoming}, \cite{nilsback2008automated}, \cite{krause20133d}, \cite{saleh2015large}, \cite{eitz2012humans}. Note that our 53-dataset covers the 8-dataset, F-CelebA and part of the Fine grained 6 tasks.} 
\label{table:53-dataset}
\vspace{-10pt}
\end{table*}


\vspace{-5pt}
\subsection{ Dataset and Baselines:} 
\vspace{-5pt}
\label{sec:4.1}

\noindent{\bf Datasets.}
We use image classification as the basic task framework. We extend the conventional benchmark 8-dataset \cite{aljundi2018memory, aljundi2017expert} to a more challenging 53-datasets by collecting more challenging classification tasks. 53-datasets consist of 53 image classification datasets. Each one supports one complex classification task, and the corresponding dataset is obtained from previously published sources, e.g., task 1 \cite{5206537}: classify scenes into 67 classes, such as kitchen, bedroom, gas station, etc (scene dataset with 15,523 images); task 2 \cite{WahCUB_200_2011}: classify 200 types of birds, such as Brewer Blackbird, Cardinal, Chuck will Widow, etc (birds dataset with 11,787 images). A full list and more details on each dataset are in supplementary materials. The 53-datasets is a subset of SKILL-102 \cite{ge2023lightweight} Lifelong Learning benchmark dataset\footnote{SKILL-102 dataset \textcolor{magenta}{\url{http://ilab.usc.edu/andy/skill102}}}, more details about dataset creation and an extended version (107 tasks) is in USC-DCT (Diverse Classification Tasks), a broader effort in our laboratory. DCT can be used for many machine vision purposes, not limited to lifelong learning. 

We use 53 datasets with $>1.8$M images from 1,584 classes over 53 tasks for experiments.
Table~\ref{table:53-dataset} shows the details of our 53-dataset in comparison to other benchmark datasets. Specifically, we compare the number of different classification targets among different benchmarks, which represents the diversity and difference among different continual tasks. For instance, our 53-dataset contains 5 different classification targets: object recognition, style classification (e.g., painting style), scene classification, counting number (e.g., CLVER dataset \cite{johnson2017clevr}), and medical diagnosis (e.g., Breast Ultrasound Dataset).
To date, our 53-dataset is one of the most challenging image classification benchmarks for continual learning algorithms, with a large number of tasks and inter-task variance.

For the experiments below, we subsampled the dataset to allow some of the sequential baselines to converge: we capped the number of classes/task to 300 (only affected 1 tasks), and used either around $5,120$ training images for tasks with $c\geq 60$ classes, or around $2,560$ for tasks with $c<60$, where $c$ represent the number of classes. Thus, we used 53 tasks, total 1,583 classes, total 132,625 training images and 13,863 test images. 
We also conduct experiments on CIFAR-100 dataset (Appendix Fig.12).
\noindent{\bf Baselines.}
As discussed in Sec~\ref{sec:introduction}, we grant each baseline a perfect task oracle during inference. We implemented 13 baselines, which can be roughly categorized in the following 3 categories \cite{de2021continual}: 

\underline{\textit{(1) Dynamic Network methods}} contains most of the baseline methods because our method also belongs to it: they dynamically modify the network to solve new tasks, usually by network expansion.
We use PSP \cite{cheung2019superposition}, Supermask in Superposition (SUPSUP) \cite{wortsman2020supermasks}, CCLL\cite{singh2020calibrating}, Confit\cite{jie2022alleviating}, and EFTs\cite{verma2021efficient} as the representative methods of this category: For PSP, the model learns all 53 tasks in sequence, generating a new PSP key for each task. The keys help segregate the tasks within the network in an attempt to minimize interference. For SUPSUP, the model uses a random initialized parameter as fixed backbone and learns class-specific supermasks for each task, which help alleviate catastrophic forgetting. During inference, different tasks use different supermasks to make the decision. 

\underline{\textit{(2) Regularization methods}} add an auxiliary loss term to the primary task objective to constraint weight updates. The extra loss can be a penalty on the parameters 
(EWC \cite{kirkpatrick2017overcoming}, LwF\cite{li2017learning}, MAS \cite{aljundi2018memory} and SI \cite{ zenke2017continual}) 
or on the feature-space (FDR \cite{benjamin2018measuring}), 
such as using Knowledge Distillation (DMC \cite{zhang2020class}). 
We use EWC, online-EWC, SI, LwF as the representatives of this category: for EWC, one agent learns all 53 tasks in sequence, using EWC machinery to constrain the weights when a new task is learned, to attempt to not destroy performance on previously learned tasks. 

\underline{\textit{(3) Replay methods}} use a buffer containing sampled training data from previous tasks, as an auxiliary to a new task's training set. The buffer can be used either at the end of the task training (iCaRL, ER \cite{rebuffi2017icarl, robins1995catastrophic}) 
or during training (GSS, AGEM, AGEM-R, DER, DERPP \cite{lopez2017gradient, chaudhry2018efficient, aljundi2019gradient, buzzega2020dark}). 
We use Episodic Memory Rehearsal (ER) as the representative baseline of this category: One agent learns all 53 tasks in sequence. After learning each task, it adds  10 images/class of that task to a growing replay buffer that will later be used to rehearse old tasks. When learning a new task, the agent learns from all the data for that task, plus rehearses all old tasks using the memory buffer. 
Additionally, SGD is a naive baseline that just fine-tunes the entire network for each task, with no attempt at minimizing interference. SGD-LL is a variant that uses a fixed backbone plus a single learnable shared last layer for all tasks with a length equal to the task with the largest number of classes (500 classes in our setting). SGD uses standard stochastic gradient descent as optimization, which may suffer from the catastrophic forgetting problem.

For a fair comparison, all the 13 baseline methods use an ImageNet-pretrained ResNet-50 \cite{he2016deep} backbone, except for PSP, Confit (requires ResNet-18) and SUPSUP (which requires a randomly initialized ResNet-50 backbone).


\begin{figure*}[ht]
\vspace{-30pt}
\centering
\includegraphics[width=\linewidth]{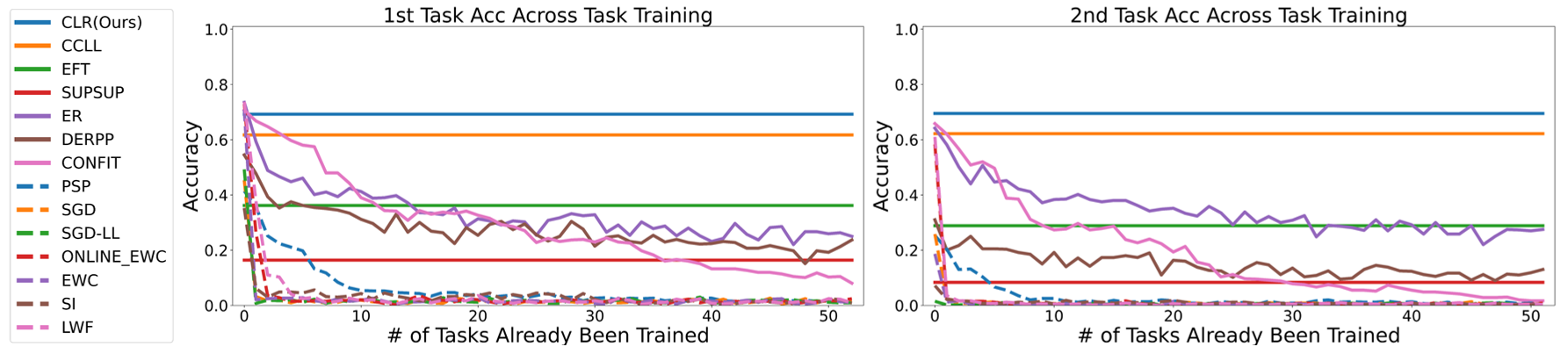}
\caption{Accuracy on task 1 as a function of the number of tasks learned. Our approach maintains the highest accuracy on task 1 over time, and importantly, it totally avoids catastrophic forgetting and maintains the same accuracy as the original training, no matter how many new tasks are learned. As discussed in the approach section, this is because we explicitly isolate the task-specific parameters for all tasks and avoid parameter interference. This is also the case for baselines SUPSUP \cite{wortsman2020supermasks}, CCLL \cite{singh2020calibrating}, and EFT \cite{verma2021efficient}. Other baseline methods suffer a different degree of catastrophic forgetting. EWC \cite{serra2018overcoming}, PSP \cite{cheung2019superposition}, LwF\cite{li2017learning}, SI \cite{ zenke2017continual} and SGD suffer severe catastrophic forgetting with this challenging dataset. Rehearsal-based method ER  performs relatively well because it has an unlimited large replay buffer, and it saves 10 images/class of the previous tasks. Yet, the overall accuracy of ER is still lower than our CLR-reprogrammed model. Rehearsal methods also incur higher (and increasing) training costs because of the rehearsing. We noticed similar performance on the second task.}
\label{fig:task1}
\end{figure*}

\begin{figure*}[th]
\centering
\includegraphics[width=.85\linewidth]{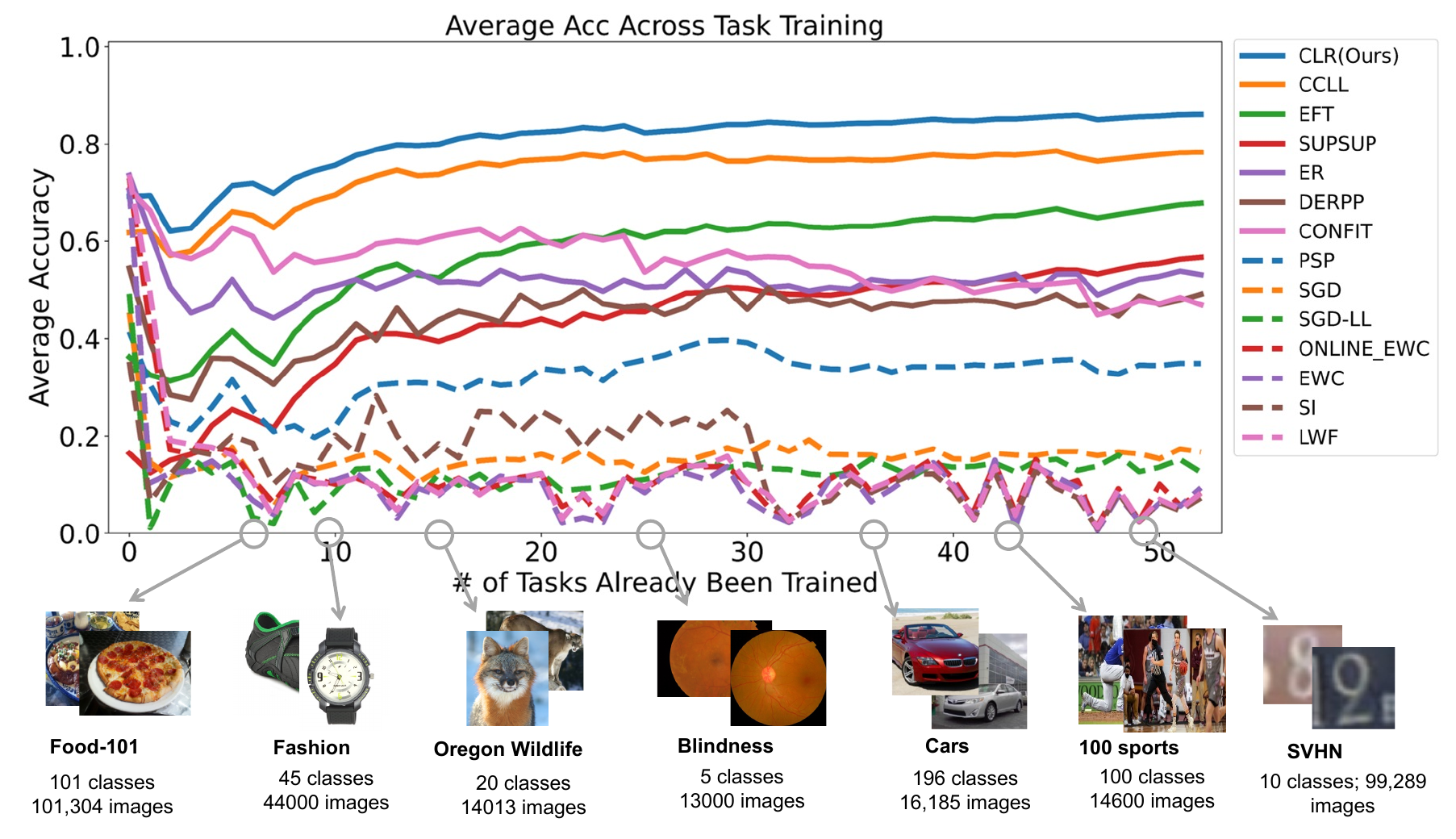}
\caption{
Average accuracy on all tasks learned so far, as a function of the number of tasks learned.  Our CLR-reprogrammed approach is able to maintain higher average accuracy than all baselines. The average accuracy increases because some  of the later tasks are easier than former tasks (i.e., later tasks have higher accuracy). }
\label{fig:avgtask}
\vspace{-10pt}
\end{figure*}

\subsection{Accuracy on the first tasks}
\label{sec:4.2}


To evaluate the performance of all methods in overcoming catastrophic forgetting, we  track the accuracy on each task after learning new tasks. If the method suffers from catastrophic forgetting, then the accuracy of the same task will decrease after learning new tasks. A great continual learning algorithm is expected to maintain the accuracy of the original learning performance after learning new tasks, which means that old tasks should be minimally  influenced by an increasing number of new tasks. Fig.~\ref{fig:task1} shows the accuracy on the first, and second tasks as we learn from 1 to 53 tasks using 53-datasets, to gauge the amount of forgetting on early tasks as many more tasks are learned.  Overall, our CLR-reprogrammed model maintains the highest accuracy on these early tasks over time, and importantly, our method (similar to CCLL \cite{singh2020calibrating}, EFT \cite{verma2021efficient} and SUPSUP \cite{wortsman2020supermasks}) avoids  forgetting and maintains the same accuracy as the original training, no matter how many new tasks are learned. SUPSUP is not able, even from the beginning, to learn task 1 as well as other methods. We attribute this to SUPSUP's limited expressivity and capacity to learn using masks over a random backbone, especially for tasks with many classes. Indeed, SUPSUP can perform very well on some other tasks, usually with a smaller number of classes (e.g., SVHN, UMNIST Face Dataset).
Baseline methods suffer a different degree of  forgetting: EWC \cite{serra2018overcoming}, PSP \cite{cheung2019superposition}, ER \cite{robins1995catastrophic}, SI \cite{zenke2017continual}, and LwF \cite{li2017learning} suffers severe catastrophic forgetting in the challenging dataset. 
We noticed similar performance on the second task.

\subsection{Average accuracy after learning all 53 tasks}
\label{sec:4.3}
\begin{figure}[t]
\vspace{-10pt}
\centering
\includegraphics[width=\linewidth]{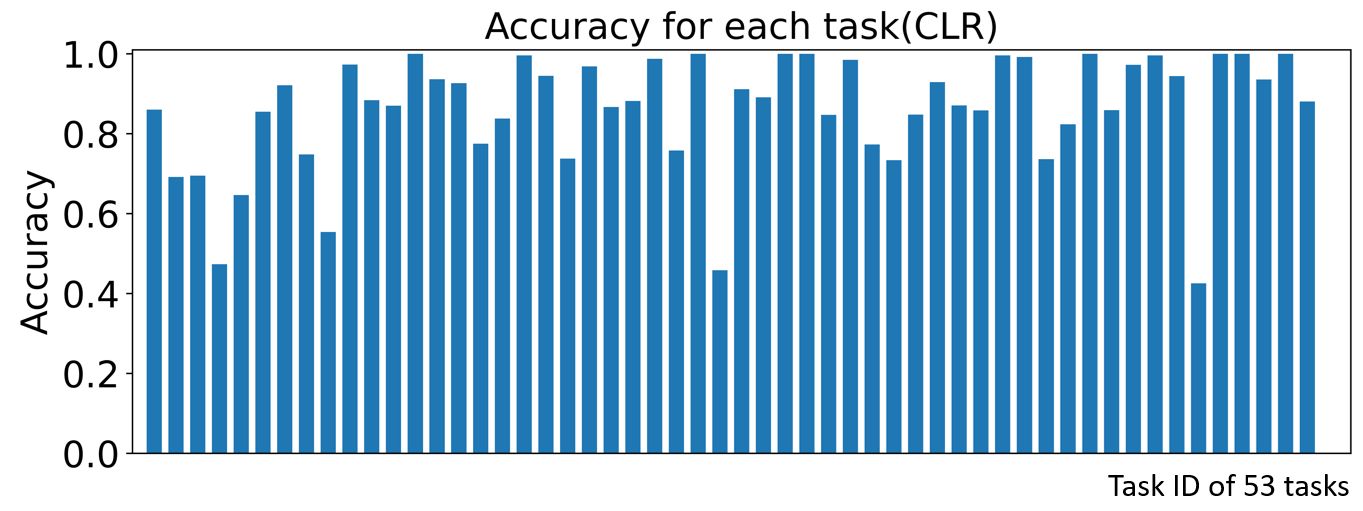}
\caption{Absolute accuracy per task after learning 53 tasks with our CLR-reprogrammed CNN.}
\label{fig:absacc}
\end{figure}

We computed the average accuracy on all tasks learned so far after learning 1, 2, 3, ... 53 tasks.
We plot the  accuracy averaged over all tasks learned so far, as a function of the number of tasks learned in Fig.~\ref{fig:avgtask}. Note that the level of difficulty for each of the 53 tasks is quite variable, as shown in Fig.~\ref{fig:absacc}. Hence, the average accuracy over all tasks so far may go up or down when a new task is added, depending on whether an easier or harder task is added. The average accuracy represents the overall performance of a continual learning method in learning and then performing sequential tasks. Overall our CLR-reprogrammed model achieves the best average accuracy compared with all baselines. 
For replay-based methods, the overall performance is lower than us even though they have a large buffer,
and the training time is increased for these methods with the increase in task number. EWC, PSP, LWF, SI suffer severe catastrophic forgetting in the challenging 53-datasets. 

To show more details, we plot the accuracy of each task after learn all 53 tasks with our CLR-reprogrammed method in Fig.~\ref{fig:absacc}. This shows that 53-dataset provides a range of difficulty levels for the various tasks, and is quite hard overall.

\subsection{Parameter and computation cost}
\label{sec:4.4}
Achieving higher overall average accuracy during learning of sequential tasks is very important. A great continual learning algorithm is also expected to minimize the requirement of extra network parameters, and the computation cost. 
Table.~\ref{table:cost} shows the required extra parameter and computation cost for our approach and the baselines. The "extra parameters to add one new task" represents the percentage compared with the original backbone (e.g., ResNet-50) size. For instance, EWC needs no extra parameter, while its computation cost is relatively high (to compute the Fisher information matrix that will guide the constraining of parameters while learning a new task), and the accuracy is not the best. Our CLR method required only 0.59\% extra parameters to learn each new task, and the computation cost increase is small compared with baseline SGD (normal training). Importantly, our method achieves the best average accuracy. 

\begin{table}[]
\begin{center}
 \small
 \rowcolors{1}{}{lightgray}
\resizebox{0.48\textwidth}{!}{
\begin{tabular}{c c c c}
\toprule
    \thead{Method}  & \thead{Extra parameter \\ to add 1 new task}  & \thead{Computation \\ cost} & \thead{Average Acc \\  (53-datasets)}\\
\midrule

SGD   & \textbf{0\%} & 1 & 16.71 \% \\
PSP \cite{cheung2019superposition}   & 5.02\% & 0.828 & 34.91 \% \\
EWC \cite{kirkpatrick2017overcoming}   & 0\% & 1.160 & 9.36 \% \\
ONLINE EWC \cite{schwarz2018progress}   & 0\% & 1.011 & $9.47^*$ \% \\
SGD-LL& \textbf{0\%} & \textbf{0.333} & 12.49 \% \\
ER \cite{robins1995catastrophic}   & 189.14\% & 3.99 & 53.13 \% \\
SI \cite{zenke2017continual}   & 0\% & 1.680 &  7.28\% \\
LwF \cite{li2017learning}   & 0\% & 1.333 & 8.23\% \\
SUPSUP \cite{wortsman2020supermasks}  & 3.06\% & 1.334 & 56.69 \% \\
EFT \cite{verma2021efficient} &    3.17\% & 1.078  & 67.8 \% \\
CCLL \cite{singh2020calibrating} &  0.62\%  & 1.006  & 78.3 \% \\
CLR (Ours) &  0.59\%  & 1.003 & \textbf{86.05 \%} \\
\bottomrule
\end{tabular}
}
\end{center}
\caption{Extra parameter expenditures and computation cost analysis. We treat the computation cost of SGD as the unit, and the computation costs of other methods are normalized by the cost of SGD. PSP's low computation cost comes from using a Resnet-18 backbone instead of Resnet-50 which is its original form. For EWC, though the final model size does not increase, the performance is poor and N fisher matrices are needed during training; EWC-online updated the way of updating the fisher matrix and only requires one fisher matrix during training. ER maintain a memory buffer that includes five images per class from the tasks that have already been seen, we spread the size in bytes of the image buffer over the 53 tasks to obtain the amount of extra parameters per task. SUPSUP requires a 3MB mask for each task.} 
\vspace{-15pt}
\label{table:cost}
\end{table}


\subsection{Influence of different immutable backbone}
\label{sec:4.5}
Our method obtains the task-agnostic immutable parameters by training the CNN model on a relatively diverse dataset with supervised learning or proxy tasks with self-supervised learning. 
To investigate the influence of different pretraining methods on the performance of continual learning, we choose four different kinds of task-agnostic immutable parameters trained with different datasets and tasks. For supervised learning, besides Imagenet-1K, we also conduct experiments with a pretrained backbone on the Pascal-VOC image classification task (relatively smaller one). For self-supervised learning, which needs no semantic labels, we conduct experiments with backbone trained with DINO \cite{caron2021emerging} and SwAV \cite{caron2020unsupervised}.
DINO \cite{caron2021emerging} is a self-distillation with no label framework, which utilizes multiple crop of the image (patch) on the same model and update model's parameters with exponential moving average. While SwAV \cite{caron2020unsupervised} simultaneously clusters the data and keeps the consistency between cluster assignments produced for different augmentations of the same image. The results in Table~\ref{table:backbone} shows that both supervised and self-supervised learning could contribute a good immutable parameters, and our method is robust to different backbone pre-training. Note how Pascal-VOC is a much smaller dataset, which may explain the lower overall accuracy; with any of the other (larger) datasets, accuracy is almost the same, suggesting that our method is not highly dependent on a specific set of backbone features.

\begin{table}[t]
\begin{center}
 \small
 \rowcolors{1}{}{lightgray}
\scalebox{.9}{%
\begin{tabular}{c c c}
\toprule
    \thead{Learning paradigm}  & \thead{Method/dataset}  & \thead{Average Acc \\  (53-datasets)}\\
\midrule

Supervised Learning   & ImageNet-1k \cite{russakovsky2014imagenet} & 86.05 \% \\
Supervised Learning   & Pascal VOC \cite{Everingham15} & 82.49 \% \\
Self-supervised Learning   & SwAV \cite{caron2020unsupervised} & 85.12 \% \\
Self-supervised Learning   & DINO \cite{caron2021emerging} & 85.77 \% \\
\bottomrule
\end{tabular}
}
\end{center}
\caption{Influence of different task-agnostic immutable parameter. Both supervised learning and self-supervised learning could contribute a relatively good immutable parameter for our method, which shows that our method is robust to different backbones.} 
\label{table:backbone}
\vspace{-5pt}
\end{table}

\section{Conclusion}
We propose Channel-wise Lightweight Reprogramming (CLR), a parameter-efficient add-on continual learning method, to allow a single network to learn potentially unlimited parallel input-to-output mappings and to switch on the fly between them at runtime. CLR adds channel-wise lightweight linear reprogramming to shift the original pretrained fixed parameter to each task, which is simple and generalizable to any CNN-based model. The experiments on continually learning 53 different and challenging tasks show that the CLR method achieves state-of-the-art performance on task incremental continual learning. Besides high performance, CLR is also parameter-efficient, which requires only 0.59\% extra parameter to learn a new task. 

\noindent{ \bf Acknowledgements}
This work was supported by Amazon ML Fellowship, C-BRIC (one of six centers in JUMP, a Semiconductor Research Corporation (SRC) program sponsored by DARPA), DARPA (HR00112190134) and the Army Research Office (W911NF2020053). The authors affirm that the views expressed herein are solely their own, and do not represent the views of the United States government or any agency
thereof.

\clearpage

{\small
\bibliographystyle{ieee_fullname}
\bibliography{main}
}

\clearpage
\section*{Appendix}
\section{Details of our 53-dataset for continual learning and method performance}
\begin{figure*}[th]
\centering
\includegraphics[width=\linewidth]{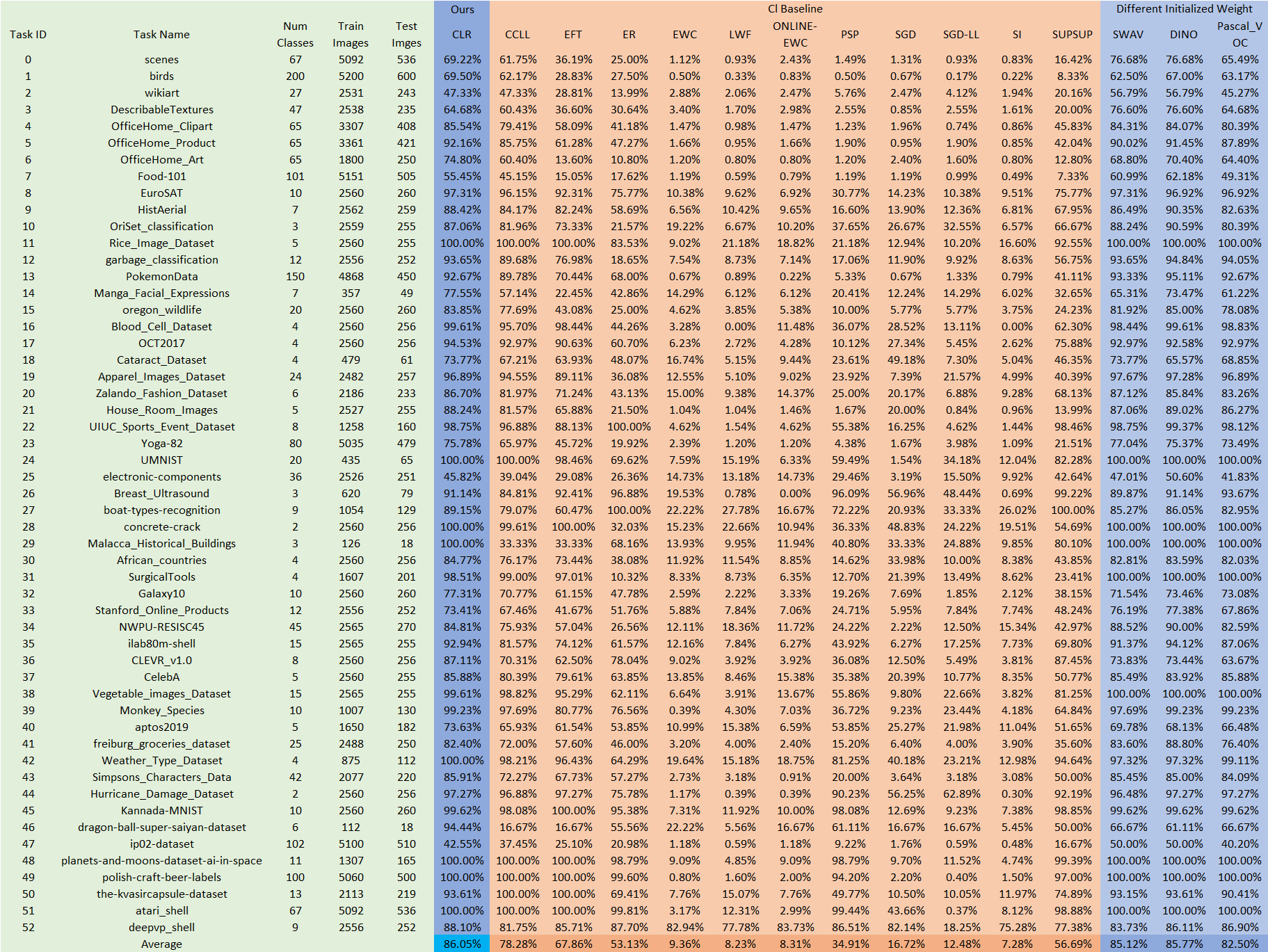}
\caption{Statistics of the datasets and per-task accuracy of our method and baselines after learning all 53 tasks in the continual learning setting. Ablation columns indicate our methods with different initialization weights.}
\label{fig:summary_result}
\end{figure*}

Fig.~\ref{fig:summary_result} shows a summary of the 53 datasets we used as the continual learning benchmark in our main paper. The figure also shows the detailed per-task accuracy of our methods and baselines after learning all 53 tasks in the task incremental continual learning setting.

\section{Channel-wise linear reprogramming ability}
To further understand the performance of channel-wise lightweight reprogramming achieved by channel-wise linear transformation,
we conduct qualitative 
experiments to explore the ability of CLR layer to transfer the feature map from a Pre-trained immutable parameter weight set (starting point) to a target parameter weight set (goal).  

Usually, the Pre-trained weight is not good enough due to the domain gap between the Pre-trained dataset/learning paradigm and the target dataset. And a relatively good performance could be achieved by either finetuning the whole backbone on the target dataset (FINETUNE) or learning from scratch (randomly initialized backbone) on the target task dataset (SCRATCH). We will show that with the help of a very cheap CLR layer, a feature map after a pretrained (non-optimal) model could be reprogrammed towards a "relatively optimal" feature map obtained by either finetuning the whole backbone (FINETUNE) or training from scratch (SCRATCH). 

We choose two datasets: CLEVR dataset and the Kannada-MNIST dataset. Model performance on the CLEVR dataset reaches 46.09\% with a Pre-trained ResNet-50 backbone + linear head, 97.66\% with FINETUNE, and 91.41\% with SCRATCH. In this scenario, pretrain has a large accuracy gap with FINETUNE and SCRATCH. It would be interesting to see if the CLR layer could reprogram a feature map obtained from pretrain towards a feature map obtained by FINETUNE and SCRATCH, which shows the ability of CLR layer to fill a large domain gap.

\begin{figure*}[th]
\centering
\includegraphics[width=\linewidth]{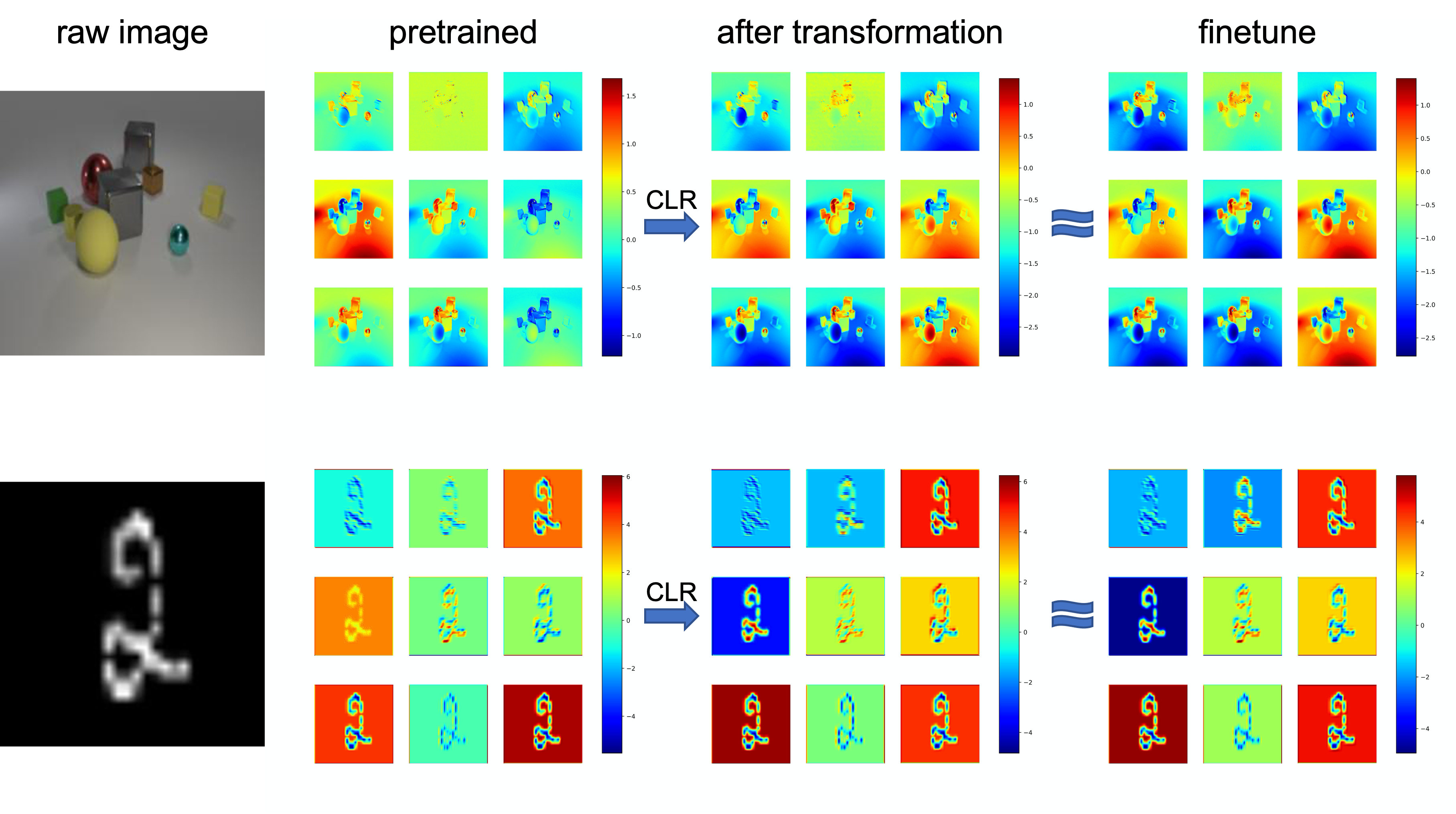}
\includegraphics[width=\linewidth]{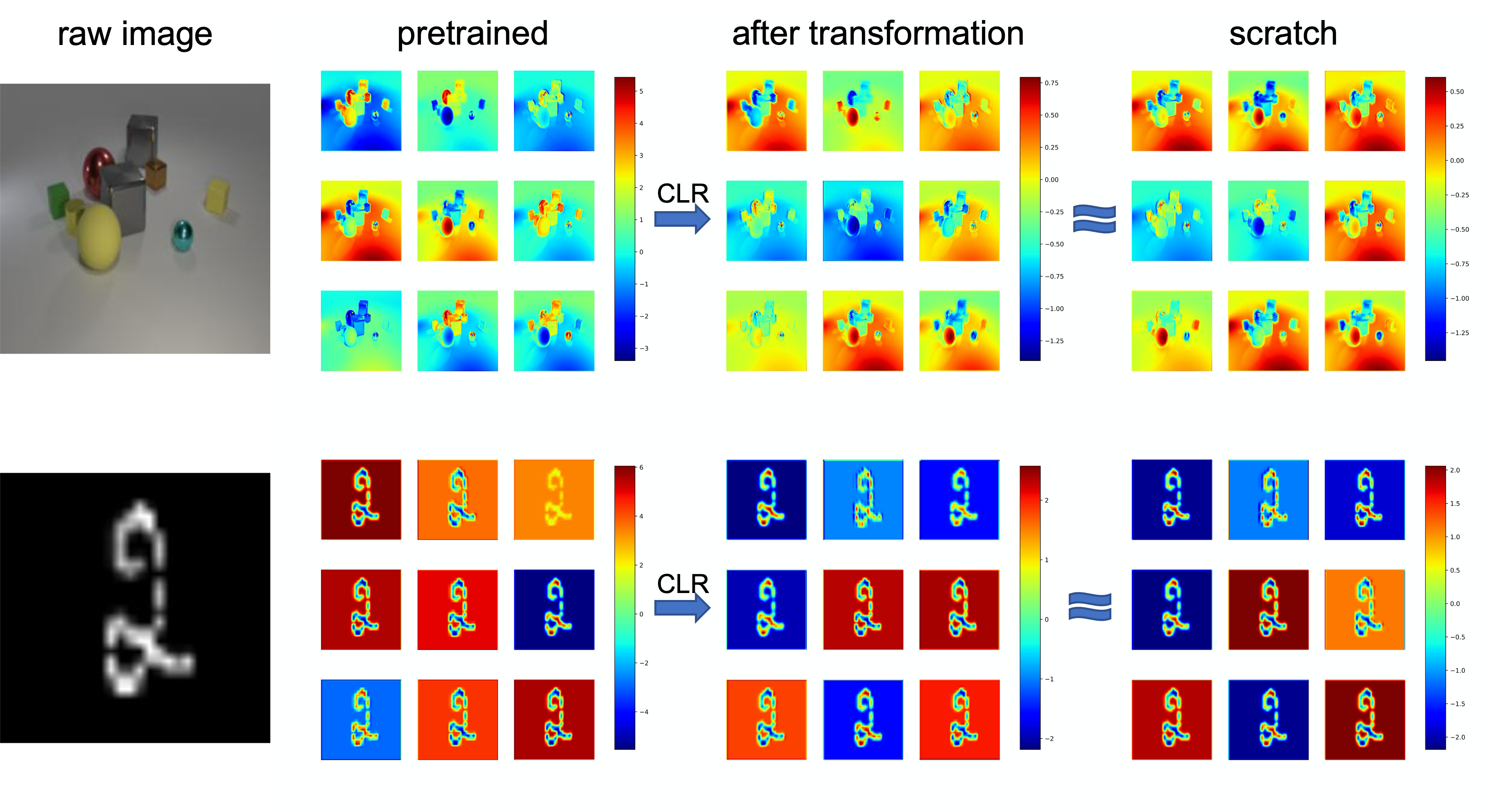}
\caption{The Figure shows quantitative results of the CLR transformation ability on CLEVR and Kannada-MNIST datasets. We visualize the feature maps in the first residual group of ResNet-50 that initially has a large initial gap between pre-train and FINETUNE (or SCRATCH). The results show that after channel-wise linear transformation, the feature after pre-train could be reprogrammed towards the goal feature (FINETUNE or SCRATCH). Pretrained indicates the frozen Imagenet pretrained ResNet-50 backbone. Finetune is a finetuned ResNet-50 backbone with Imagenet pretrained initialization while Scratch is a trained ResNet-50 backbone with random initialization.}
\label{fig:visual}
\end{figure*}

Model performance on the Kannada-MNIST dataset reaches 95.77\% with a Pre-trained backbone + linear head, 99.62\% with FINETUNE, and 100\% with SCRATCH. Here, SCRATCH performs better than FINETUNE, which shows that the pretrained weights may have no benefit (or even harmful) for target task learning. Here we want to show that the CLR layer could reprogram a feature map obtained from pretrain towards the feature map obtained by SCRATCH. We use the feature map after the first convolutional layer in the different models (pretrain, FINETUNE, and SCRATCH). Taking the feature map after the pretrain model as input and the feature map after FINETUNE (or SCRATCH) as output, we
utilize a CLR layer (3x3 2D depthwise convolutional kernels) to learn the mapping, i.e. the channel-wise linear transformation between them. 
The qualitative results are shown in Fig.~\ref{fig:visual}.
Specifically, in Fig.~\ref{fig:visual}, we visualize the feature map that initially has a large initial gap between pretrain and FINETUNE (or SCRATCH). The results show that after channel-wise linear transformation, the feature after pretrain could be reprogrammed towards the goal feature (FINETUNE or SCRATCH)

\section{Bootstrapping results}

Fig.5 in the main paper shows the average accuracy as more tasks are learned. However, the gradient of the curve is also influenced by the order of the tasks (i.e., Hard tasks located in earlier sequence will cause average accuracy tends to increase,
while easy tasks located in earlier sequence will cause average accuracy tends to decrease) which is entangled with the effect of catastrophic forgetting. 


\begin{figure*}[th]
\centering
\includegraphics[width=\linewidth]{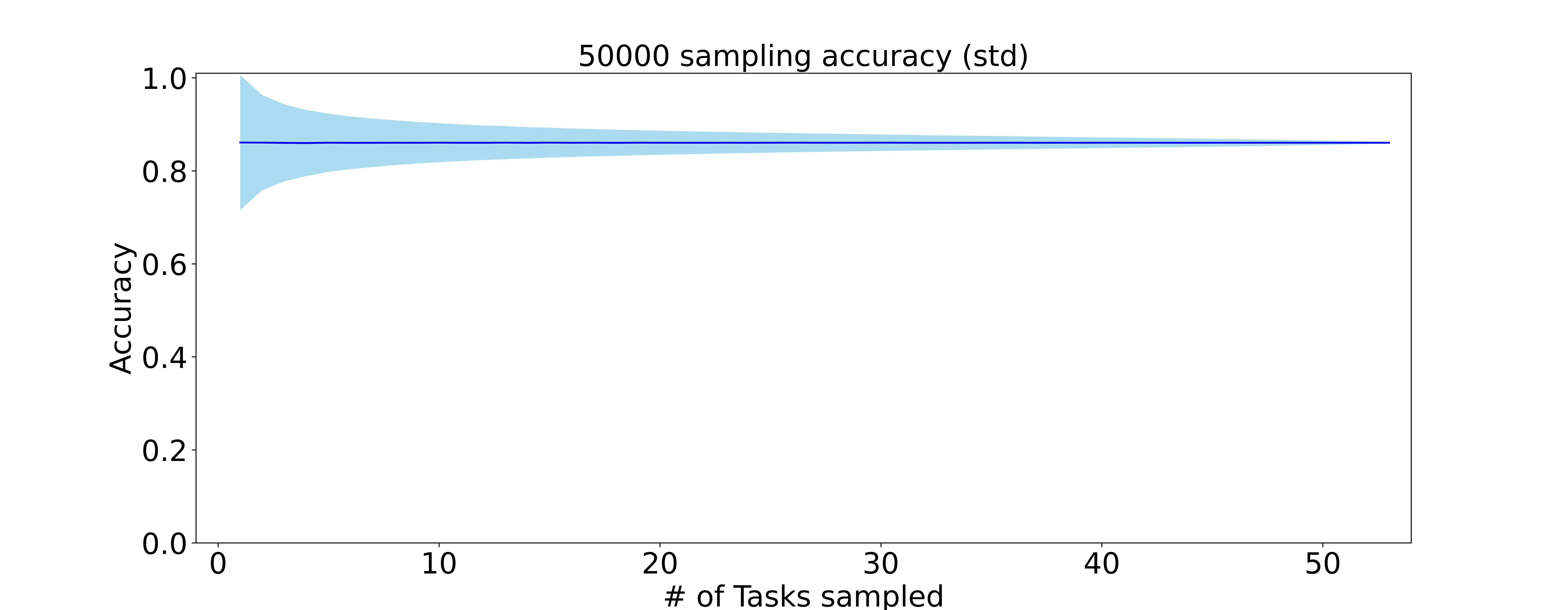}
\caption{Bootstrapping statistic results. The X-axis represents the number of tasks $t$ in a specific continual learning task. Y-axis shows the mean Accuracy (solid blue line) on the sampled tasks with replacement and std as the shaded light blue range.}
\label{fig:bootstrapping-std}
\end{figure*}

\begin{figure*}[th]
\includegraphics[width=\linewidth]{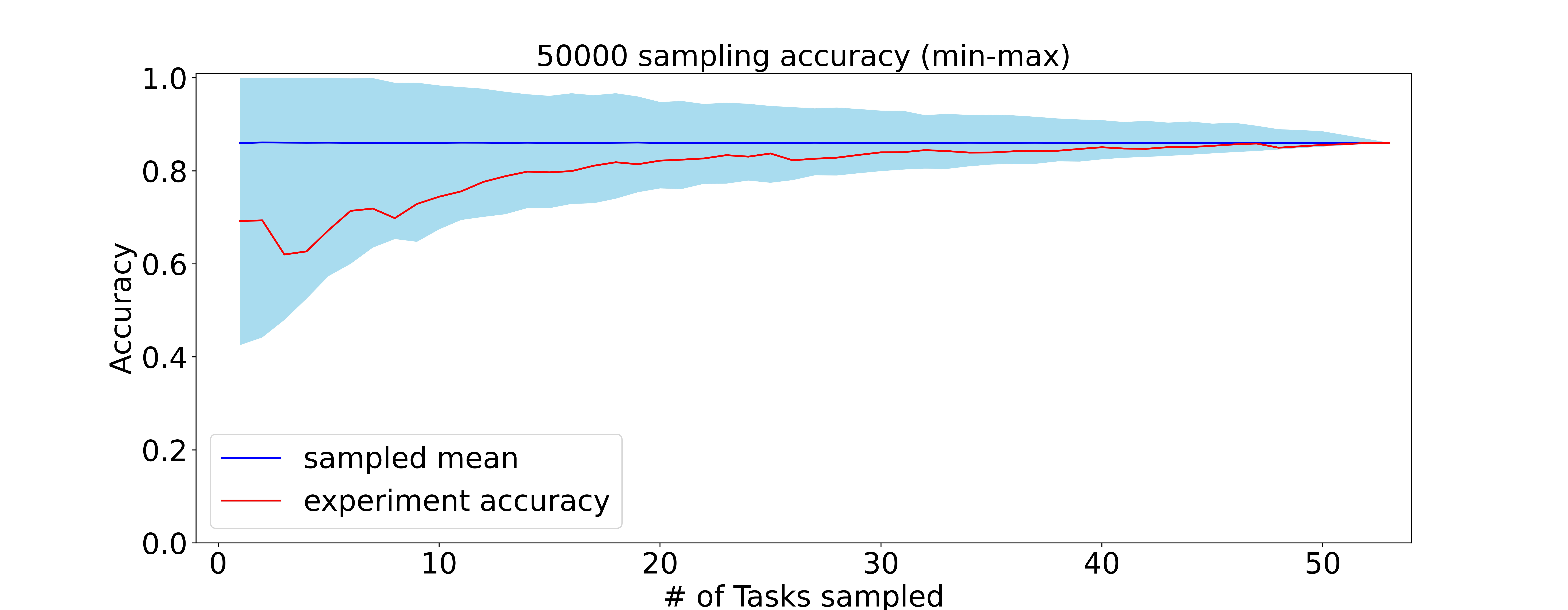}
\caption{Bootstrapping statistic results with detailed accuracy log. The X-axis represents the number of tasks $t$ in a specific continual learning task. Y-axis shows the mean Accuracy (solid blue line) on the sampled tasks with replacement. The shaded light blue range shows the min and max range in the given task number $t$. We use the solid red line to represent our reported results in the main paper Fig.5, which filled in the shaded light blue range.}
\label{fig:bootstrapping-minmax}
\end{figure*}



We use bootstrapping to show the tendency of average accuracy when more tasks are learned. Specifically, for any number of tasks ($t\in (1,...,53)$) that we want to conduct in one continual learning setting, we randomly sample $t$ tasks from the 53 tasks 50,000 times with replacement and compute the mean accuracy (mean) and standard deviation (std). Fig.~\ref{fig:bootstrapping-std} shows the Bootstrapping statistic results, which show the change of mean and std when we increase the total number of tasks. The X-axis represents the task number $t$ we want to conduct. For instance, if the continual learning task number $t$=10, then we randomly sample 10 tasks from the 53-dataset and calculate the mean accuracy. We repeat the sampling 50000 times and get the std. Y-axis shows the mean Accuracy (solid blue line) on the sampled tasks with replacement and std as the shaded light blue range.
Since in our CLR method, the order of task is not mattered (we have the same performance on a specific task no matter the sequence), this allows us to simulate what would happen if we learn a different sample of tasks given a specific task number $t$. We observe that the mean accuracy is stable and not influenced by $t$ when the sample number is large.
For std, when the task number $t$ is small, the std is relatively large, and the std decreases with task number $t$ increase. When $t$=53,  the std becomes zero. 

Fig.~\ref{fig:bootstrapping-minmax} shows the Bootstrapping statistic results with detailed max and minimum accuracy logs. The X-axis represents the number of tasks $t$ in a specific continual learning task. Y-axis shows the mean Accuracy (solid blue line) on the sampled tasks with replacement. The shaded light blue range shows the min and max range in the given task number $t$ among 50000 times task samples. We use the solid red line to represent our reported results in the main paper Fig.5, which filled in the shaded light blue range.

\section{More experiments to explore the trade-off between parameter and performance}
Several other versions of our method may include methods with higher accuracy but higher cost. Our main method - CLR (the one in the paper) adds the CLR layer after all the original convolutional kernels except for the 1$\times$1 kernels, saving many parameters. 

CLR-Full version applying the CLR layer to all convolutional kernels in the pretrained model which reaches 85.85\% in accuracy and cost 1.69$\times$ parameters compared to our main method (CLR). 

The CLR-Reduced version adds a smaller version of CLR layer with 1$\times$1 2D reprogramming kernels after all 1$\times$1 original Conv kernels and normal CLR layer with 3$\times$3 2D reprogramming kernels after the rest CONV kernels. It reaches 85.7\% in accuracy and costs 1.08$\times$ parameters compared to our main method (CLR). 

The CLR-mixed version learns a weighted combination of the original and our reprogrammed feature maps. The intuition is that we keep some proportion of the original feature and add the new features learned after reprogramming. Specifically, 
A trainable parameter \textbf{A} decides the weight of the summation of the reprogrammed feature and the original feature map. Equation~\ref{eq:1} shows the details of the weighted combination.
\begin{equation}
    \hat{x'}_k = A*CLR_{k}(x'_k) + (1-A)*x'_k
\label{eq:1}
\end{equation}
where $x'_k$ is the kth channel of the feature map from the original Convolutional layer and $CLR_{k}$ is the corresponding linear transformation. It reaches 86.25\% in accuracy and costs 1.79$\times$ parameters compared to our main method (CLR). 

\begin{figure*}[th]
\centering
\includegraphics[width=\linewidth]{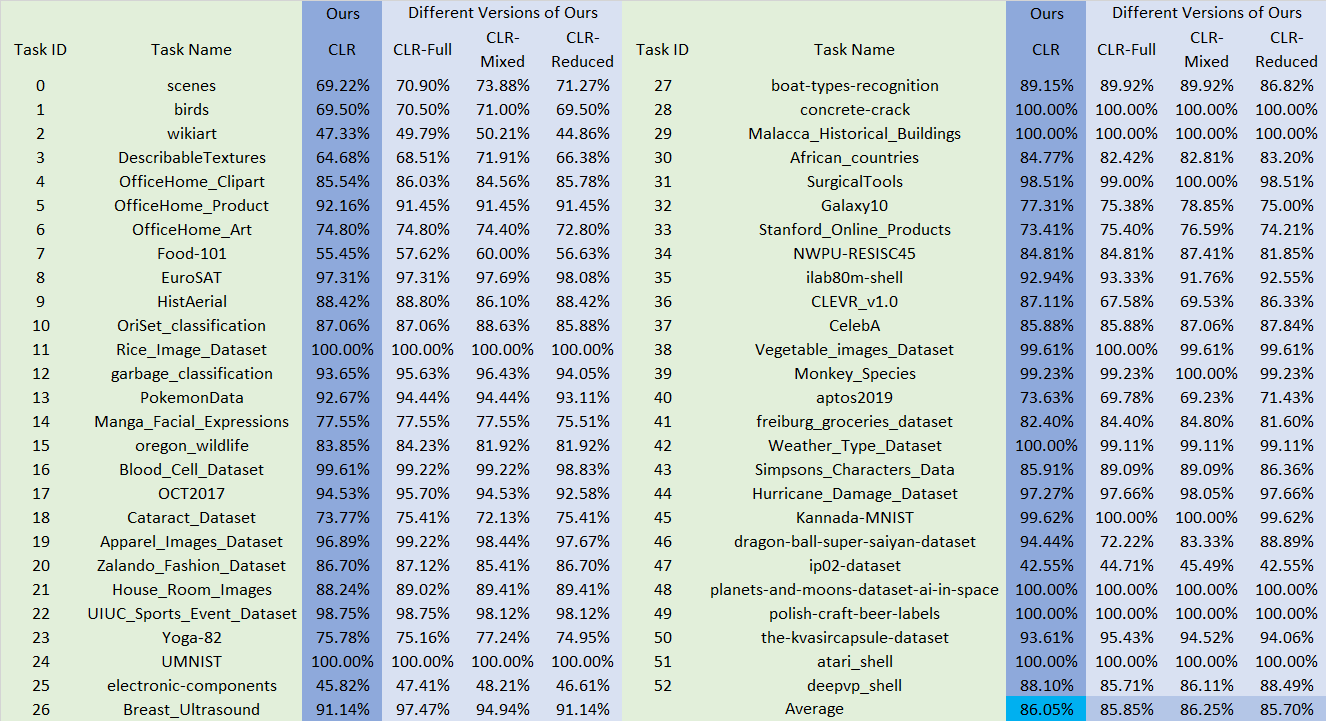}
\caption{Per-task accuracy of our main method and other versions of our method after learning all 53 tasks in the continual learning setting.}
\label{fig:ablation}
\end{figure*}

The results are shown in Fig.~\ref{fig:ablation}. Theoretically, more trainable parameters could lead to better performance, and it is true for CLR-mixed version, which achieves +0.2\% than our main method. Interestingly, the CLR-full version achieves lower average accuracy than the main method, while most of the per-task accuracy is higher (43 out of 53 tasks). 

\section{Transfer learning with CLR-based model}
\label{sec:4.2}
We apply our CLR method to the transfer learning problem, where we only care about the accuracy of the transferred dataset while do not need to maintain performance on previous datasets.

\noindent{\bf Datasets.} We use the same 53-dataset to evaluate transfer learning performance. Specifically,  we use the  ImageNet pretrained ResNet-50 model as initialization and apply our method and 4 baseline  transfer learning methods 53 times, on 53 different classification tasks.

\noindent{\bf Baselines.} We have four baseline methods: 1) learn from scratch (SCRATCH), where the backbone ResNet-50 is randomly initialized with no prior knowledge, and then uses the training set of each task to train the whole network from scratch. 2) finetune the whole backbone and last layer (FINETUNE), 3) finetune only the last layer (LINEAR), 4) Head2Toe method \cite{evci2022head2toe} use the fixed backbone and need two steps: 1) feature selection: train the model by adding a large fully connection between all intermediate features and the last layer and select the important connection by adding regularization. 2) keep the important skip connection and retrain the added layers.

\begin{figure*}[th]
\centering
\includegraphics[width=\linewidth]{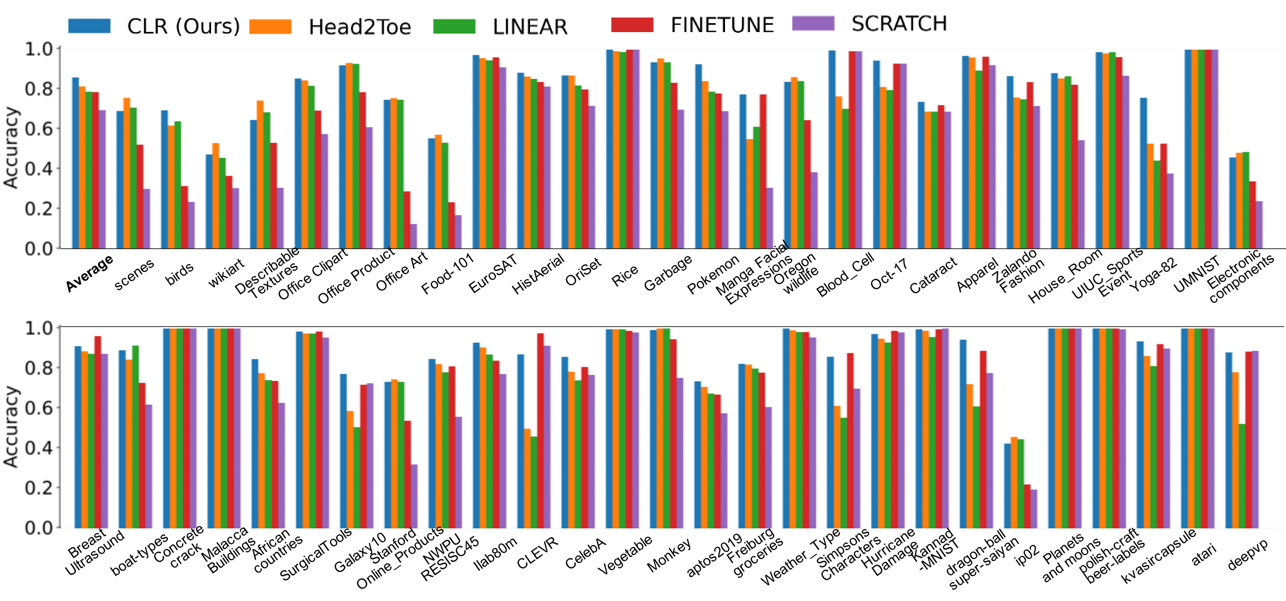}
\caption{Bar plot of transfer learning performance on 53-dataset.}
\label{fig:transfer}
\end{figure*}

\begin{figure*}[th]
\centering
\vspace{-0.2cm}
\includegraphics[width=\linewidth]{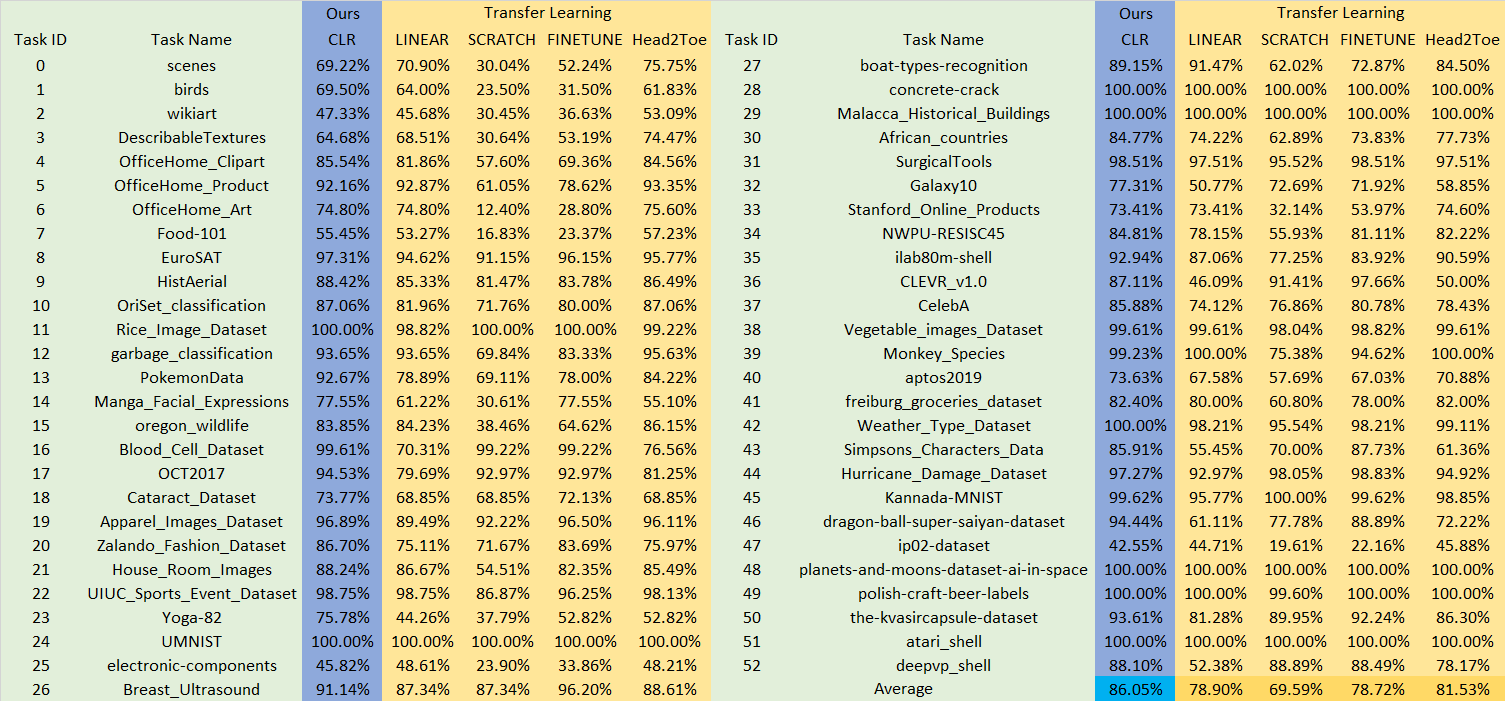}
\caption{Transfer learning result on 53-dataset of our method and other baselines (LINEAR, SCRATCH, FINETUNE, and Head2Toe)}
\label{fig:transfer_result}
\end{figure*}

Fig.~\ref{fig:transfer} shows the average accuracy on all 53 classification tasks and the details of each task, and Fig.~\ref{fig:transfer_result} shows the detailed result for transfer learning. Our CLR achieves the best average accuracy on the 53-dataset compared with all baselines. Specifically, CLR achieves almost 5\% average improvement on 53 datasets over Head2Toe, and even larger improvement over LINEAR, FINETUNE, and SCRATCH. This shows the effectiveness of the CLR-based model in transfer learning problems.

\section{CIFAR-100 Result}
We also show our method's result on incremental CIFAR-100 dataset with other previous baselines in table \ref{table:cifar}

\begin{table}[]
\begin{center}
 \small
 \rowcolors{1}{}{lightgray}
\resizebox{0.2\textwidth}{!}{
\begin{tabular}{c c}
\toprule
    \thead{Method}  & \thead{Average Acc}\\
\midrule
LwF & 24\% \\
iCARL & 49\% \\
RPS & 57\% \\
CCLL & 85\% \\
EWC & 41\% \\
SI & 52\% \\
CLR (Ours) &  \textbf{94.2\%} \\
\bottomrule
\end{tabular}
}
\end{center}
\caption{We applied our method on CIFAR-100 dataset with 10 tasks, each containing 10 classes with comparisons to baselines from CCLL, using ResNet-18 as the backbone.} 
\vspace{-15pt}
\label{table:cifar}
\end{table}

\end{document}